\pdfoutput=1

\documentclass[11pt]{article}

\usepackage[preprint]{acl}

\usepackage{times}
\usepackage{latexsym}

\usepackage{comment}
\usepackage{booktabs}
\usepackage{multirow}

\usepackage[T1]{fontenc}

\usepackage[utf8]{inputenc}

\usepackage{microtype}

\usepackage{inconsolata}

\usepackage{algorithm}
\usepackage{algpseudocode}
\usepackage{amsmath}
\usepackage{tcolorbox}

\newcommand{\luis}[1]{\textcolor{red}{#1}}

\title{\textsc{CHEW}: A Dataset of \textsc{CH}anging \textsc{E}vents in \textsc{W}ikipedia}

\author{
\textbf{Hsuvas Borkakoty\textsuperscript{1}},
\textbf{Luis Espinosa-Anke\textsuperscript{1,2}},
\\
\textsuperscript{1}Cardiff NLP, School of Computer Science and Informatics, Cardiff University, UK 
\\
\textsuperscript{2}AMPLYFI, UK
\\
\small{
   \texttt{\{borkakotyh,espinosaankel\}@cardiff.ac.uk}
 }
}

\begin{document}
\maketitle
\begin{abstract}
We introduce \textsc{CHEW}, a novel dataset of changing events in Wikipedia expressed in naturally occurring text. We use \textsc{CHEW} for probing LLMs for their ``timeline'' understanding of Wikipedia entities and events in generative and classification experiments. Our results suggest that LLMs, despite having temporal information available, struggle to construct accurate timelines. We further show the usefulness of \textsc{CHEW}-derived embeddings for identifying meaning shift.
\end{abstract}

\section{Introduction}

Since language models (LMs) are trained on raw web text and, often, without any explicit temporal grounding \cite{zhao2024set}, they are prone to suffer \textit{temporal misalignment} \cite{luu2021time,lazaridou2021mind,jang2022temporalwiki}. While there is a significant body of work concerned with fixing this issue via, e.g., in-domain pretraining \cite{gururangan2020don}, neologism-focused pretraining \cite{yu2021dict}, knowledge editing \cite{de2021editing,zhu2020modifying,dai2021knowledge}, continual learning \citep{agarwal2021temporal,del2018short,giulianelli2020analysing,dhingra2022time,loureiro2022timelms}, or model editing \cite{rosin2022temporal}, there is however little understanding on LLMs' ability to reproduce timelines of entities and events, primarily due to pretraining \textit{temporal chaos} \cite{zhao2024set}.


Probing models for representing temporal knowledge has been the focus of prior works, e.g., in the lexical semantics space, where studying semantic change or meaning shift has been used as a proxy to explore internal representations of word meaning, typically via diachronic embeddings \cite{del2019short, schlechtweg2020semeval, loureiro2022tempowic} or generating temporally grounded definitions \cite{giulianelli2023interpretable, luden2024beyond}. Moreover, ``time-sliced'' perplexity has examples ranging from Wikipedia to Twitter \cite{cheng2024dated,loureiro2022timelms}. Conversely, temporal question answering (tasks where the correct answers change over time) probe factual and world knowledge on LLMs with some kind of time context \cite{liska2022streamingqa,kasai2024realtime,zhao2024set,wallat2024temporal}. Despite this, 
there are not enough evaluation benchmarks to probe for change modeling, with TemporalWiki \cite{jang2022temporalwiki} partly addressing this issue by curating a Wikipedia \textit{diffs} dataset, although the authors themselves admit it is not trivial to tell if a change in Wikipedia or Wikidata content signify meaningful world changes. In this paper, we deep dive on the notion of change by proposing \textsc{CHEW} (\textsc{CH}anging \textsc{E}vents in \textsc{W}ikipedia), a temporally grounded dataset from Wikipedia that focuses on finding important changes to events and entities, starting from 
a collection of Wikipedia events and entities, and their associated changes over time extracted from Wikipedia lists and originally curated in the TAQA dataset \cite{zhao2024set}. We report generation, classification and downstream results using CHEW, shedding light on LLMs' capabilities to handle temporal information in various settings, and their potential for temporal alignment. 

\begin{table*}[!t]
\LARGE
\resizebox{0.95\textwidth}{!}{
\begin{tabular}{@{}rrp{8.5cm}lp{8.5cm}r@{}}
\toprule
\multicolumn{1}{c}{\textbf{Wikipedia Title}} & \multicolumn{1}{c}{\textbf{Timestamp 1}} & \multicolumn{1}{c}{\textbf{Text 1}} & \multicolumn{1}{c}{\textbf{Timestamp 2}} & \multicolumn{1}{c}{\textbf{Text 2}} & \multicolumn{1}{c}{\textbf{Label}} \\ \midrule
 Andrés Iniesta  & 05-01-2017 & Andrés Iniesta Luján (born 11 May 1984) is a Spanish professional footballer who plays as a central midfielder for FC Barcelona and the Spain national team. He serves as the captain for Barcelona... & 27-12-2018 & Andrés Iniesta Luján (born 11 May 1984) is a Spanish professional footballer who plays as a central midfielder for Japanese club Vissel Kobe... & change    \\ \midrule
Sonotone & 30-12-2009 & The Sonotone 1010 hearing aid, introduced December 1952, was the was the first commercial product to use transistors ...& 29-12-2010 & The Sonotone 1010 hearing aid, introduced on 29 December 1952, was the first commercial product to use transistors ...  & no change \\ \bottomrule
\end{tabular}
}
\caption{Examples of \textsc{CHEW} which illustrate the significant changes in entities contained in the positive examples, as opposed to stylistic differences in negative examples.}
\label{tab:examples}
\end{table*}

\section{Building \textsc{CHEW}}






The TAQA dataset \cite{zhao2024set}, denoted as \( D_{\text{TAQA}} \), comprises Wikipedia articles with temporal question-answer pairs. Our goal is to derive from it \( C = P \cup N \), a set of Wikipedia page pairs representing temporal changes \( P \) and their corresponding negative examples \( N \). Let \( Q \) be the set of all questions and \( T \) the set of all timestamps. For a question \( q \in Q \) and timestamps \( t_1, t_2 \in T \) where \( t_1 < t_2 \), let \( A(q, t) \) denote the answer to question \( q \) at time \( t \). We define \( \Delta A(q, t_1, t_2) = A(q, t_1) \neq A(q, t_2) \) to identify changes in answers. From \( D_{\text{TAQA}} \), we extract pairs \( (q, (t_1, t_2)) \) such that \( \Delta A(q, t_1, t_2) \) holds. For each valid pair, we obtain the revisions of the corresponding Wikipedia articles at \( t_1 \) and \( t_2 \). Specifically, consider the ranked list of revisions at \( t_1 \), \( \{R_{t_1}^1, R_{t_1}^2, \ldots, R_{t_1}^n\} \), sorted by their timestamps in ascending order. We select the first revision from this list, denoted as \( R_1 = R_{t_1}^1 \), and similarly, the first revision from the list at \( t_2 \), denoted as \( R_2 = R_{t_2}^1 \). We then compute cosine similarities using SBERT \cite{reimers2019sentence}, specifically \( S(R_1, R_2) \), \( S(A(q, t_1), A(q, t_2)) \), and \( S(A(q, t_2), R_1) \). Pairs satisfying \( S \geq \theta \) for all three similarities\footnote{We empirically set \(\theta\) to 0.6.} are retained, and form \( P \). This filtering step ensures the following criteria are met:  (1) $R_1$ and $R_2$ reflect the change available in \( D_{\text{TAQA}} \), which was generated from a manually curated Wikipedia list, and therefore is accurate\footnote{Note that relying on Wikipedia lists has other advantages, since Wikipedia topics are popular and well structured, and their distribution is less biased than open knowledge graphs \cite{piscopo2017wikidata}.}; (2) $R_1$ and $R_2$ are sufficiently similar, which ensures to a great extent that no other important changes have occurred between $t_1$ and $t_2$; and (3) We can guarantee that the evidence for the change that originated that positive datapoint is contained in one sentence (as opposed to, e.g., split across several sentences using anaphoric references). 

\begin{figure}[!t]
    \centering
    \begin{tabular}{c}
        \includegraphics[width=\columnwidth]{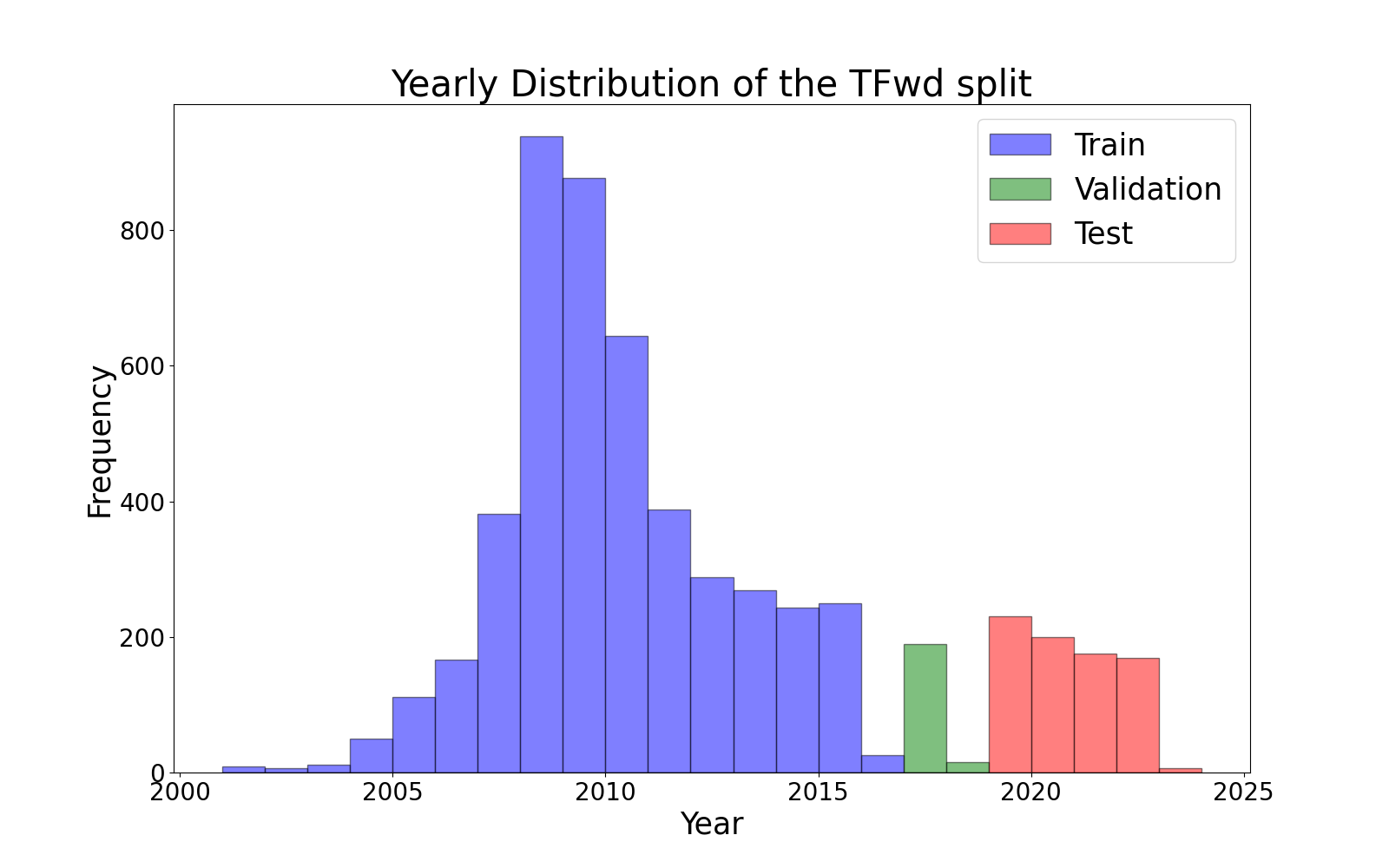} \\
        \text{(a) Time Forward Split (TFwd).} \\
        \includegraphics[width=\columnwidth]{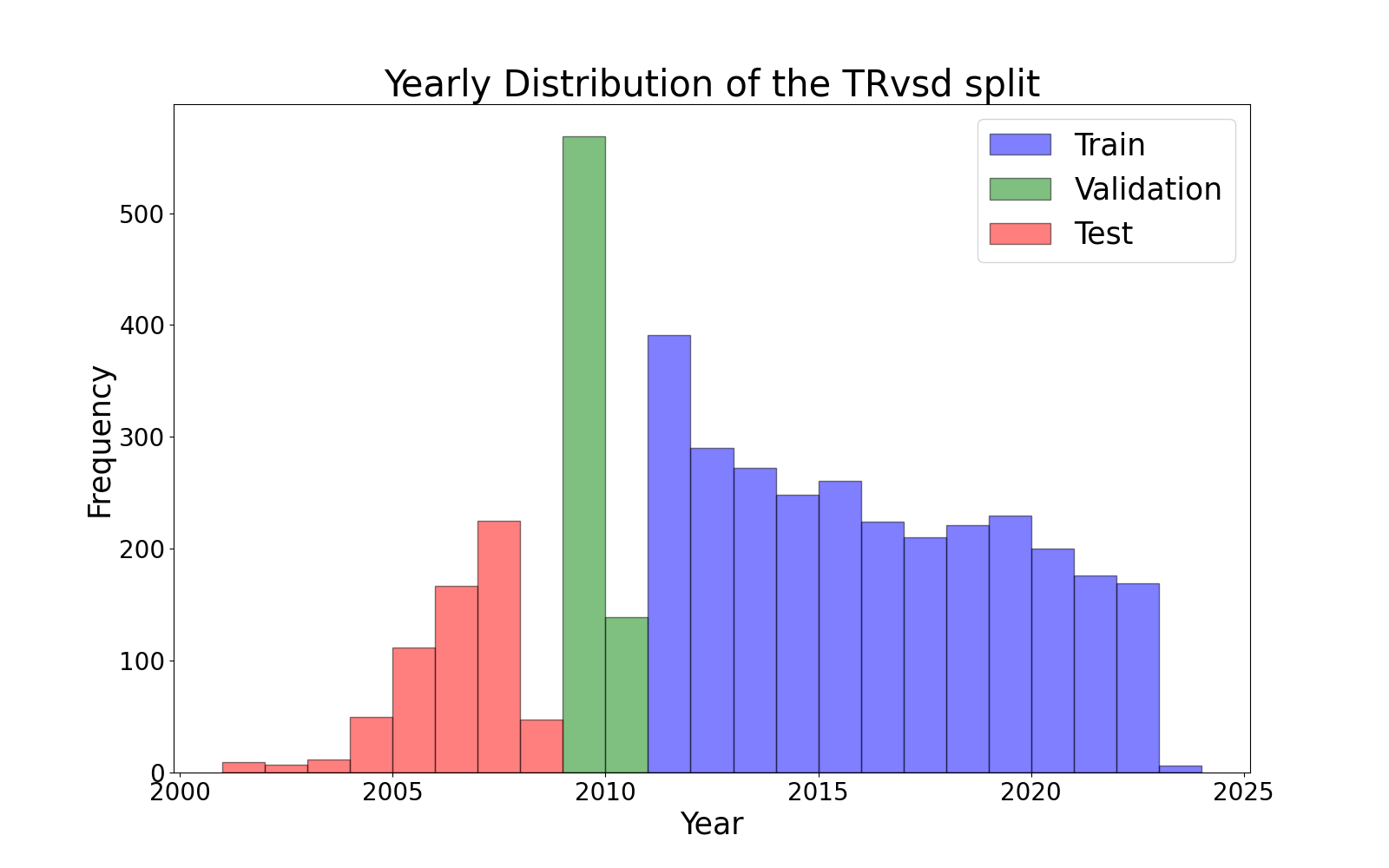} \\
        \text{(b) Time Reversed Split (TRvsd).} \\
    \end{tabular}
    \caption{Barplots with the time forward and time forward data splits.}
    \label{fig:time-splits}
\end{figure}

For curating \( N \) (i.e., no change), we simply sample the first and last snapshot \( R_{t_1} \) and \( R_{t_2} \) of a period in the timeline of a Wikipedia page that was highly edited. We then filter out those pairs where the cosine similarity of their definition sentences, \( S(R_{t_1}^{\text{def}}, R_{t_2}^{\text{def}}) \), is sufficiently high, but not 1 (again, empirically, between 0.8 and 1). After manual inspection of a sample, we found that this step ensures that the negative examples represent pairs of Wikipedia pages that are different, but with no significant change. Statistics on \textsc{CHEW} are provided in Table \ref{tab:stats}, for several splits: \textbf{Random}, where we randomly separate all pairs between predefined train/validation/test splits; \textbf{No overlap} between entities (NoOv); \textbf{Time-forward} (TFwd) and \textbf{Time-reversed} (TRvsd), the latter two for no temporal overlap across splits (see Fig. Figure \ref{fig:time-splits}).

\begin{table}[!t]
\scriptsize
\resizebox{\columnwidth}{!}{
\begin{tabular}{@{}l|lrrr@{}}
\toprule
\multicolumn{1}{c|}{\textbf{Data split}} & \multicolumn{1}{c}{\textbf{Set}} & \multicolumn{1}{c}{\textbf{No ch.}} & \multicolumn{1}{c}{\textbf{Change}} & \multicolumn{1}{c}{\textbf{Total}} \\ \midrule
\multirow{3}{*}{Random}                  & Train                            & 1,632                                  & 2,649                               & 4,281                              \\
                                         & Val                              & 470                                    & 141                                 & 611                                \\
                                         & Test                             & 942                                    & 282                                 & 1,224                              \\ \midrule
\multirow{3}{*}{NoOv}              & Train                            & 2,106                                  & 2,193                               & 4,299                              \\
                                         & Val                              & 307                                    & 301                                 & 608                                \\
                                         & Test                             & 631                                    & 578                                 & 1,209                              \\ \midrule
\multirow{3}{*}{TFwd}            & Train                            & 2,828                                  & 2,076                               & 4,904                              \\
                                         & Val                              & 90                                     & 341                                 & 431                                \\
                                         & Test                             & 126                                    & 655                                 & 781                                \\ \midrule
\multirow{3}{*}{TRvsd}           & Train                            & 1,498                                  & 2,283                               & 3,781                              \\
                                         & Val                              & 1,407                                  & 299                                 & 1,706                              \\
                                         & Test                             & 139                                    & 490                                 & 629                                \\ \bottomrule
\end{tabular}
}
\caption{\textsc{CHEW} statistics for the four splits we introduce.}
\label{tab:stats}
\end{table}

\section{Experiments}
\label{sec:prompting}

\begin{figure}[!t]
\centering
\begin{tabular}{c}
\includegraphics[width=1\columnwidth]{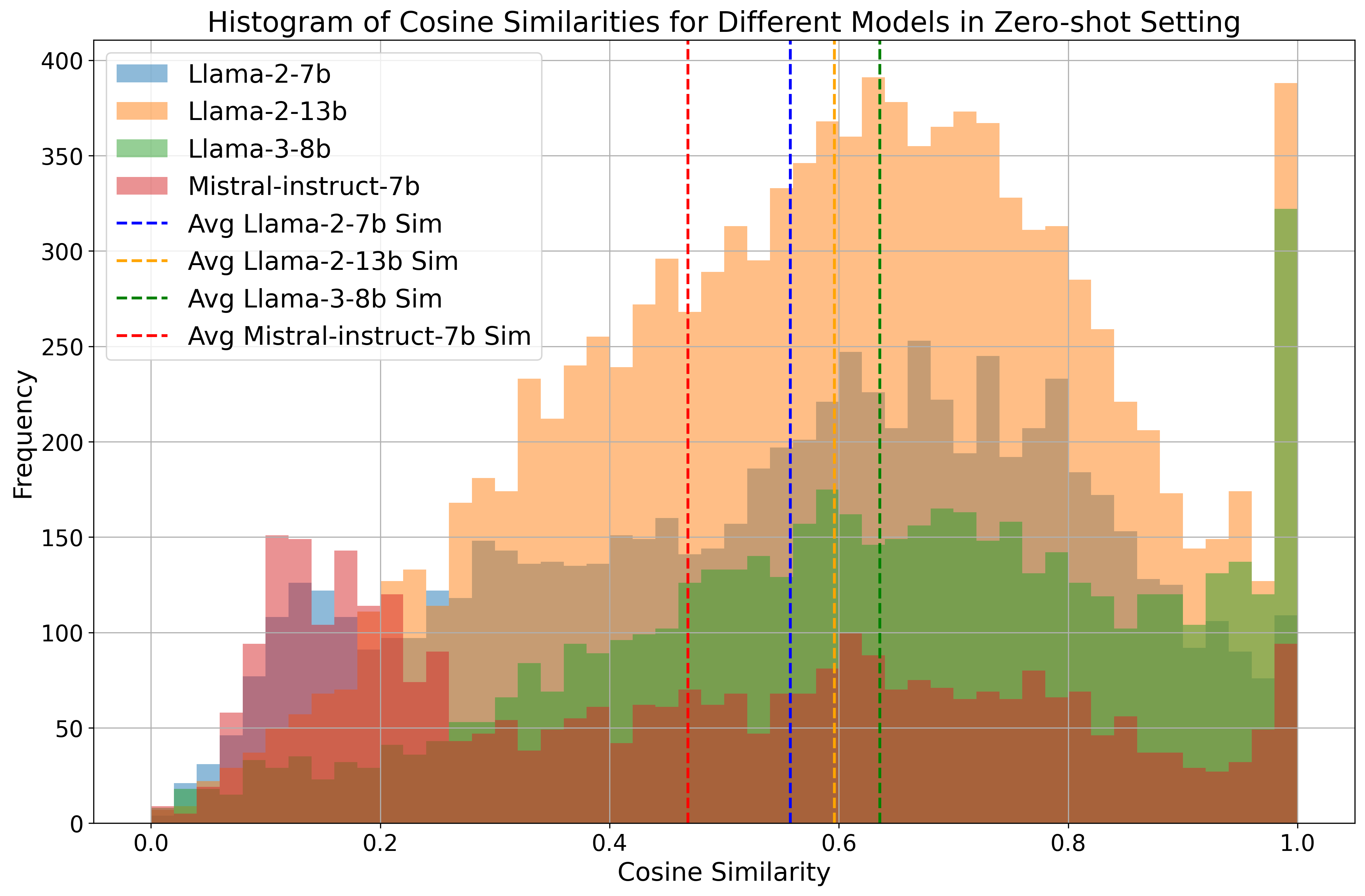} 
\end{tabular}
\caption{Similarities comparing ground truth and generations after probing for temporal knowledge.}
\label{fig:histograms}
\end{figure}

\paragraph{\noindent \textbf{Prompting for timeline knowledge}} 
Given timestamps \( t_1 \) and \( t_2 \), we submit to an LLM \( LLM \) a prompt \( p \) with a tuple \( (i, w_{t1}, t_2) \), where \( i \) is the instruction (further details can be found in Appendix \ref{app:llm_changes}), \( w_{t1} \) is the revision of \( w \) at timestamp \( t_1 \), and \( t_2 \), where \( t_1 < t_2 \). 
We then estimate the accuracy of the LLMs' response $LLM(i, w_{t1}, t_2)$ by retrieving the maximum SBERT text similarity \( S \) between the response and the content in \( w_{t2} \) as follows:

\[
S = \max \{ \text{SBERT}(r, w_{t2}) \mid r \in LLM(i, w_{t1}, t_2) \}
\]

where \( \text{SBERT}(r, w_{t2}) \) is the cosine similarity between the response \( r \) and the content in \( w_{t2} \). The higher the similarity score, the more accurate the LLM's response was. For this and subsequent experiments, we consider the following popular open source models: LLama2-7B and LLama2-13B \cite{touvron2023llama2}, LLama3-8B \cite{touvron2023llama} and Mistral-7B \cite{jiang2023mistral}. After collecting the responses and performing some ad-hoc cleanup, we visualize the similarities obtained for all models in Figure \ref{fig:histograms}. 
The most capable model in this setup is LLam2-13B, whereas Mistral-7B seems to be struggling the most to generate appropriate responses. 
An interesting observation is that Llama3-8b and Llama2-13b are actually capable to reproduce \textit{verbatim} the content of $w_{2}$, as we can see from the large number of comparisons with cosine similarity being 1. This is a somewhat surprising fact, and suggests that these models are in principle capable to generate a reasonable portion of timestamped Wikipedia articles. 

\begin{table*}[!t]
\scriptsize
\resizebox{0.98\textwidth}{!}{
\begin{tabular}{@{}ll|rrr|rrr|rrr|rrr@{}}
\toprule
                           &               & \multicolumn{3}{c|}{CHEW-Random}                               & \multicolumn{3}{c|}{CHEW-NoOv}                                 & \multicolumn{3}{c|}{CHEW-TFwd}                                 & \multicolumn{3}{c}{CHEW-TRvsd}                                \\ \midrule
                           &               & \multicolumn{1}{c}{P} & \multicolumn{1}{c}{R} & \multicolumn{1}{c|}{F1} & \multicolumn{1}{c}{P} & \multicolumn{1}{c}{R} & \multicolumn{1}{c|}{F1} & \multicolumn{1}{c}{P} & \multicolumn{1}{c}{R} & \multicolumn{1}{c|}{F1} & \multicolumn{1}{c}{P} & \multicolumn{1}{c}{R} & \multicolumn{1}{c}{F1} \\ \midrule
\multirow{4}{*}{\textbf{n.i.}} & llama2-7b     & 0.82                  & 0.82                 & 0.82                 & 0.65                  & 0.62                  & 0.61                  & 0.53                  & 0.54                 & 0.53                  & 0.61                  & 0.51                 & 0.55                  \\
                           & llama2-13b    & 0.81                  & 0.81                  & 0.81                 & 0.59                 & 0.55                 & 0.48                  & 0.50                  & 0.49                 & 0.49                 & 0.41                  & 0.50                 & 0.45                 \\
                           & llama3-8b     & 0.87                  & 0.88                 & \underline{{0.87}}                   & 0.66                  & 0.66                 & \underline{{0.66}}                  & 0.55                  & 0.57                 & \underline{0.56}                  & 0.57                  & 0.50                 & 0.53                 \\
                           & mistral-7b & 0.66                  & 0.65                 & 0.65                 & 0.59                  & 0.59                  & 0.59                  & 0.52                  & 0.52                 & 0.52                 & 0.63                  & 0.57                 & \underline{0.59}                   \\ \midrule
\multirow{4}{*}{\textbf{i.}}    & llama2-7b     & 0.81                  & 0.81                  & 0.81                 & 0.77                  & 0.77                  & \textbf{\underline{0.78}}                 & 0.52                  & 0.53                 & 0.52                  & 0.55                  & 0.51                 & 0.53                  \\
                           & llama2-13b    & 0.93                  & 0.93                  & \underline{\textbf{0.93}}                  & 0.77                  & 0.77                  & \textbf{\underline{0.78}}                 & 0.46                  & 0.46                 & 0.46                  & 0.64                  & 0.52                 & 0.57                  \\
                           & llama3-8b     & 0.87                  & 0.87                 & 0.87                 & 0.78                  & 0.79                  & \textbf{\underline{0.78}}                  & 0.56                  & 0.58                 & 0.57                  & 0.42                  & 0.50                 & 0.45                  \\
                           & mistral-7b & 0.93                  & 0.93                  & \underline{\textbf{0.93}}                 & 0.69                  & 0.69                  & 0.69                  & 0.61                  & 0.70                 & \textbf{\underline{0.65}}                  & 0.75                  & 0.64                 & \textbf{\underline{0.69}}                  \\ \bottomrule
\end{tabular}
}
\caption{SFT fine-tuning results on the different CHEW splits, with and without temporal system prompt instruction. Best F1 scores per split are highglighted in bold, best results within the specific instruction setting are underlined.}
\label{tab:mainresults}
\end{table*}

\begin{figure}[!t]
\centering
\begin{tabular}{c}
\includegraphics[width=0.9\columnwidth]{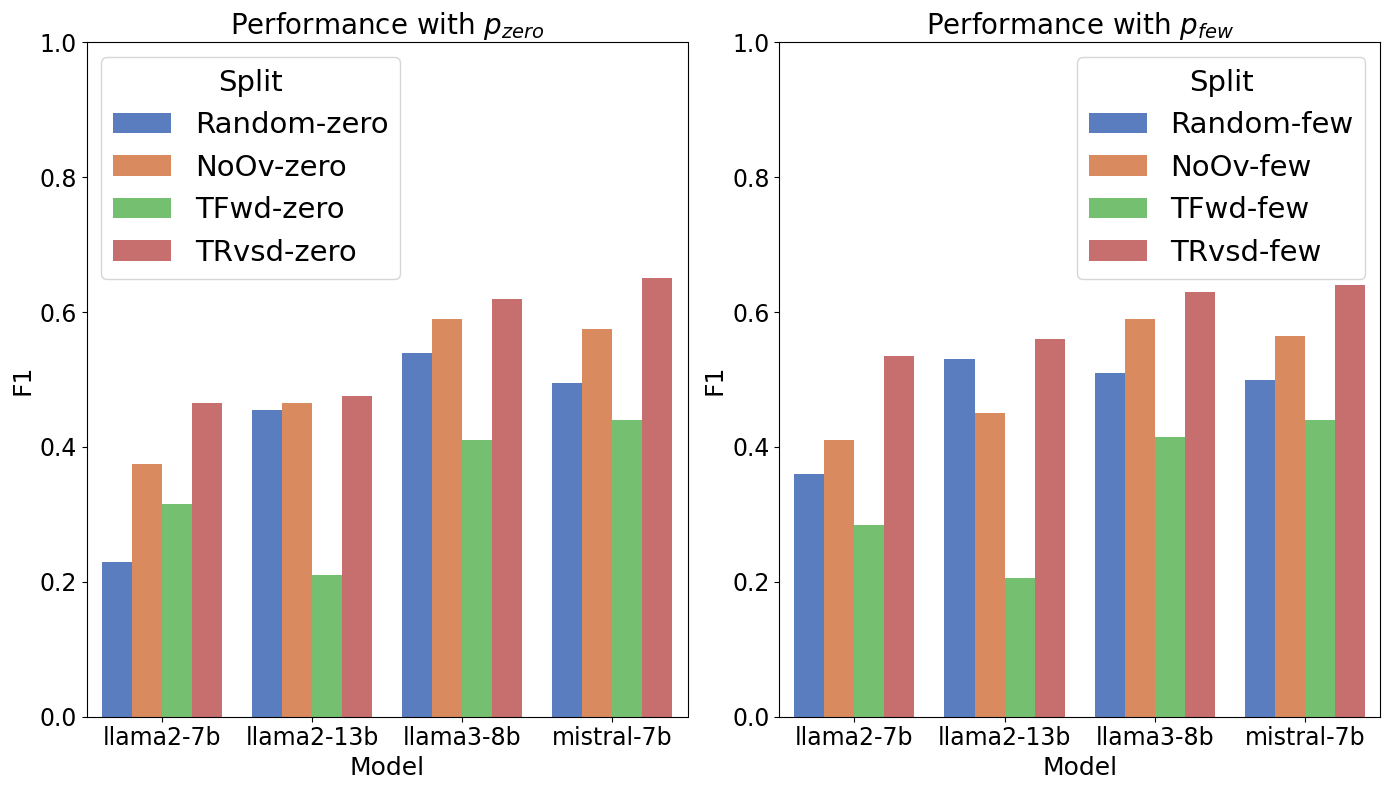}
\end{tabular}
\caption{Prompt-based classification change prediction results.}
\label{fig:promptclf}

\end{figure}

\paragraph{\noindent \textbf{Prompt-based change detection}} 
We now explore a complementary dimension of this probing experiment: binary text classification. Here, we are still in the non parametric space, and we simply prompt LLMs for a binary label, given the pair $\left(w_1, w_2\right)$. Figure \ref{fig:promptclf} shows the results of this experiment. We report results on the test sets of all splits (simply to have a point of comparison with the supervised approaches we will discuss later), and just like in the previous experiment, we report results of ``bare'' prompting ($p_{\text{zero}}$) as well as with in-context examples ($p_{\text{few}}$) (see Appendix \ref{app:prompt-based-changes} for details of these prompts and the examples provided). Our results yield two immediate observations. First, all models struggle significantly more when prompted with newer entities, and on the other hand, Llama-2 seems to struggle more than the others across the board in all splits. Interestingly, while struggling to accurately generate new information, Mistral is in fact often the best model in this classification task, which motivates us to explore it further in a downstream task (cf. Section \ref{sec:downstream}).

\paragraph{\noindent \textbf{Fine-tuning experiments}}

We proceed to fine-tune the considered models using standard LoRa \cite{hu2021lora} and SFT \cite{ouyang2022training} techniques on the training sets (Appendix \ref{sec:appendix:finetuningimpl} for implementation details). The results provided in Table \ref{tab:mainresults} yield a number of interesting insights. Note that we study two different system prompts: one where we simply refer to the task as a binary classification problem (\textbf{n.i}), and one where the models are given more context about the task (\textbf{i.}). 
Mistral-7b, which was already showing signs of being capable of handling this task, benefits substantially from the fine-tuning strategy, outperforming the Llama models in three out of four splits. Moreover, we find that Llama3-8b is only the best of the Llama models in 2 out of 4 settings, which highlights the potential of the older Llama-2 models to enhance their temporal knowledge via fine-tuning.

\section{Better temporal embeddings with CHEW}
\label{sec:downstream}

Enhancing an LLM's temporal capabilities is one of the long-standing goals of creating temporal datasets. 
We find that TempoWiC \cite{loureiro2022tempowic} is a suitable benchmark to test whether this is the case. 
This is a binary classification problem where given a pair of \textit{timestamped} tweets and a target word, models must determine whether the meaning of the target word has changed in context to the tweets. This idea closely aligns with our proposed approach of detecting changes, and so we test whether LLMs pre-trained on \textsc{CHEW} improve over base models. The upper bound for the test set in this dataset is around 
77\% macro-F1 \cite{lyu2022mllabs}, and 50\% for a random baseline. 

Specifically, we generate embeddings with both base and CHEW-finetuned LLMs, and use them in a number of ways: as features for a logistic regression classifier (contextualized embeddings concatenated and averaged), as well as deriving cosine similarities, which are then themselves the features for the logistic regression classifier (following \citet{loureiro2022tempowic}'s official baselines). We consider different layers (last, last 4, or all layers), and we select Mistral-7b base and \textsc{CHEW}-finetuned as our target model, since it showed to benefit the most from the finetuning step. The results clearly indicate an improvement on the quality of the embeddings after \textsc{CHEW}-finetuning, making this model on par with encoder-only baselines based on RoBERTa, which are known to be much better than decoder-only models for text representations 
\cite{Tang2019UnderstandingNM,behnamghader2024llm2vec}.



\begin{table}[!h]
\resizebox{\columnwidth}{!}{%
\begin{tabular}{@{}llrrrr@{}}
\toprule
\multicolumn{1}{c}{\textbf{Training data}} & \multicolumn{1}{c}{\textbf{Experiment}} & \multicolumn{1}{c}{\textbf{Acc}} & \multicolumn{1}{c}{\textbf{P}} & \multicolumn{1}{c}{\textbf{R}} & \multicolumn{1}{c}{\textbf{F1}} \\ \midrule
\multirow{5}{*}{CHEW+TW}                   & Similarity-last                         & 0.68                             & 0.69                           & 0.68                           & \textbf{0.68}                            \\
                                           & Similarity-last4                        & 0.67                             & 0.7                            & 0.67                           & \textbf{0.67}                            \\
                                           & Similarity-all                          & 0.63                             & 0.41                           & 0.63                           & 0.5                             \\ \cmidrule(l){2-6} 
                                           & Concat                                  & 0.6                              & 0.4                            & 0.63                           & 0.49                            \\
                                           & Averaged                                & 0.6                              & 0.5                            & 0.6                            & \textbf{0.54}                            \\ \midrule
\multirow{5}{*}{TW}                        & Similarity-last                         & 0.67                             & 0.58                           & 0.66                           & 0.61                            \\
                                           & Similarity-last4                        & 0.68                             & 0.55                           & 0.65                           & 0.59                            \\
                                           & Similarity-all                          & 0.58                             & 0.5                            & 0.57                           & 0.5                             \\ \cmidrule(l){2-6} 
                                           & Concat                                  & 0.56                             & 0.49                           & 0.56                           & 0.49                            \\
                                           & Averaged                                & 0.48                             & 0.5                            & 0.49                           & 0.49                            \\ \bottomrule
\end{tabular}
}
\caption{TempoWiC Results.}
\label{tab:tempowic-mistral}
\end{table}

\section{Conclusions and Future Work}
\label{sec:conclusion}

We have introduced \textsc{CHEW}, a dataset of Changing Events in Wikipedia, with which we hope to contribute to research in continual learning, temporal alingment, and other areas involving LLMs and their ability (or lack thereof) to make sense of temporal information at various levels, old and new, seen and not seen during pretraining, etc. 
We showed that aligning LLMs to temporal change makes them surprisingly competitive in the downstream task of word-in-context temporal classification when compared with their base counterparts.

\section{Limitations}
\label{sec:limitations}
Our work does not extensively evaluate a wider range of LLMs. We also have made assumptions in the similarity comparisons between Wikipedia revisions, and while our automatic, manual and downstream tests are all consistent, further extending the comparisons between revisions could lead to a more accurate dataset. Moreover, we have only focused on the English Wikipedia, which is a significant limitation, especially for exploring ``tail'' entities.



\section{Ethics statement}
\label{sec:ethics}

We believe that updating LLMs' knoweldge over time, given their prevalence and how high the interest on Generative AI is today, is critical for deploying accurate and trustworthy AI models. Therefore, it should be noted that flagging critically new content in Wikipedia might conflict with Wikipedia quality standards, as well as raise misinformation concerns (if, for example, false information peppers a Wikipedia page in a way that a change detection model is unable to capture). Further research into ensuring highly accurate scans for change of community resources remains critical today more than ever, again, especially due to how prevalent GenAI tools are today.

\bibliography{custom}

\newpage

\appendix

{\small{
\section{Prompt for Generation of changes}
\label{app:llm_changes}
The instruction prompt that we use to generate the change from a LLM is given below:

\begin{tcolorbox}
\texttt{You are a helpful knowledge management expert, and you excel at identifying critical and fundamental changes in Wikipedia entities.
You will be given an Wikipedia entity: \{ENTITYNAME\} with text in timestamp1: \{TIMESTAMP1\}, and another timestamp timestamp2:\{TIMESTAMP2\} where there has been some critical change(s) to the entity in between \{TIMESTAMP1\} and \{TIMESTAMP2\}.
Your task is to identify the changes that happened between \{TIMESTAMP1\} and \{TIMESTAMP2\} for the \{ENTITYNAME\}. Only give me the most critical information. The input contains the following:}

\texttt{
    \begin{itemize}
        \item entity: The name of the entity
        \item timestamp1: The timestamp of the text, which is provided
        \item text1: The article text of the entity in timestamp1
        \item timestamp2: The timestamp for which you have to identify the changes
    \end{itemize}
}

\texttt{Your output must be in the format of a JSON with the \{ENTITYNAME\} as its key and python list of changes as value, formatted as follows: }

\noindent \texttt{OUTPUT:}
\begin{verbatim}
{
    ENTITYNAME: python list of changes
    
}
\end{verbatim}

\texttt{If nothing really meaningful happened to \{ENTITYNAME\} during that timespan, return an empty JSON. You must return only the JSON as the output. Do not return anything else except the JSON.}

\end{tcolorbox}
\section{Prompt-based change detection experiment}
\label{app:prompt-based-changes}
The instruction prompt that we use to generate the Change Label from a LLM is as follows.

\begin{tcolorbox}
\texttt{
You are a helpful knowledge management expert and a binary classifier, and you excel at identifying critical and fundamental changes in Wikipedia entities.
You will be given an Wikipedia entity: \{ENTITYNAME\} with text in two different versions or in two different timestamps where there has been some critical change(s) to the entity in between the texts.
Your task is to identify the changes that happened between two texts of the \{ENTITYNAME\} and provide a label.\\
}
\texttt{
The input contains the following:
\begin{itemize}
    \item entity: The name of the entity
    \item text1: The first revision text of the article in the timestamp given with the text. 
    \item text2: The second revision text of the article in the timestamp given with the text.
\end{itemize}
}
\texttt{Both text1 and text2 will be formatted as follows:}\\

\texttt{<t> entity name </t> <y> timestamp </y> article text.}\\

\texttt{Your output must be in the format of a JSON with two keys label, formatted as follows: }

\noindent \texttt{OUTPUT:
    0 or 1
Where 
}
\texttt{
\begin{itemize}
    \item Label: 0 if there is no critical change in the entity between the texts
    \item Label: 1 if there is a critical change. A syntactic or grammatical change is not a critical change, only information change/update is considered a critical change.
\end{itemize}
}
\texttt{
You must return only the label, nothing else except the label(0 or 1).
}

\end{tcolorbox}

The in-context examples along with this prompt used for Few-shot learning are as follows.

1. \textbf{Positive Example:}
\begin{tcolorbox}
INPUT:\\
    \{
        \texttt{'entity': 'Blake Harrison',}\\
        \\
        \texttt{'text1': '<t> Blake Harrison </t> <y>2018-01-13T23:28:45Z </y> Blake Harrison (born 1985) is an English actor, best known for playing Neil Sutherland in the BAFTA-winning E4 comedy The Inbetweeners. Blake starred in three series and two subsequent films of the multi-award-winning comedy The Inbetweeners. Harrison's other television work includes the BBC Three comedies Way to Go and Him \& Her, Comedy Central's Big Bad World, The Bleak Old Shop of Stuff, and The Bill. Harrison also starred in all three seasons of The Increasingly Poor Decisions of Todd Margaret, created by David Cross.',}\\
        \texttt{'text2': '<t> Blake Harrison </t> <y>2019-12-19T15:57:21Z </y> Blake Harrison (born 22 July 1985) is an English actor and dancer.  Harrison starred in three series and two subsequent films of the multi-award-winning comedy The Inbetweeners. Harrison's other television work includes the BBC Three comedies Way to Go and Him \& Her, Comedy Central's Big Bad World, The Bleak Old Shop of Stuff, and The Bill. Harrison also starred in all three seasons of The Increasingly Poor Decisions of Todd Margaret, created by David Cross. '}
        \}

    \texttt{OUTPUT:\\1}
\end{tcolorbox}

2. \textbf{Negative Example:}
\begin{tcolorbox}
\texttt{INPUT:}
    \{
        \texttt{'entity': 'Miss Virginia USA',}\\
        \\
        \texttt{'text1': '<t> Miss Virginia USA </t> <y>2019-01-23T03:45:45Z </y> The Miss Virginia USA competition is the pageant that selects the representative for the state of Virginia in the Miss USA pageant. Virginia has been only moderately successful in terms of number of semi-finalists. They have had two Miss USAs. They are one of only four states to have had two Miss USAs in succession (the others being Illinois, Texas, and District of Columbia).',}\\
        \texttt{'text2':  <t> Miss Virginia USA </t> <y>2020-12-17T23:48:27Z </y> The Miss Virginia USA competition is the pageant that selects the representative for the state of Virginia in the Miss USA pageant. Virginia has been only moderately successful in terms of number of semi-finalists. They have had two Miss USAs. They are one of only four states to have had two Miss USAs in succession (the others being Illinois, Texas, and District of Columbia). '}
        \}

    \texttt{OUTPUT:\\ 0}   
\end{tcolorbox}

\section{Models and training details}
\label{sec:appendix:finetuningimpl}
We use the chat/instruct version of the models from Huggingface in our experiment, fine-tuning them using LoRA\cite{hu2021lora}. The model is loaded in 4-bit and for the task `SEQ\_CLS'(Sequence Classification). We train the models using one NVIDIA A100 GPU and inference using one NVIDIA RTX4090, taking approximately 2 hours per epoch. The training details for the models are listed below.

\begin{itemize}
    \item Learning Rate: 2e-6
    \item Optimizer: paged\_adamw\_8bit
    \item Batch size (train/eval): 1 
\end{itemize}}

The list of models and their huggingface repository names are listed in Table \ref{tab:model-repos}.

\begin{table}[]
\resizebox{\columnwidth}{!}{%
\begin{tabular}{ll}
\hline
\textbf{Model Name} & \textbf{Huggingface Repository}     \\ \hline
llama2-7b           & meta-llama/Llama-2-7b-chat-hf       \\
llama2-13b          & meta-llama/Llama-2-13b-chat-hf      \\
llama3-8b           & meta-llama/Meta-Llama-3-8B-Instruct \\
mistral-7b          & mistralai/Mistral-7B-Instruct-v0.3  \\ \hline
\end{tabular}%
}
\caption{List of Models used in our experiments and their huggingface repositories.}
\label{tab:model-repos}
\end{table}

}

\end{document}